Article

# Improving user specifications for robot behavior through active preference learning: Framework and evaluation

Nils Wilde[1], Alexandru Blidaru[1], Stephen L Smith[1] and Dana Kulić[1,2]

**Abstract**
*An important challenge in human–robot interaction (HRI) is enabling non-expert users to specify complex tasks for autonomous robots. Recently, active preference learning has been applied in HRI to interactively shape a robot's behavior. We study a framework where users specify constraints on allowable robot movements on a graphical interface, yielding a robot task specification. However, users may not be able to accurately assess the impact of such constraints on the performance of a robot. Thus, we revise the specification by iteratively presenting users with alternative solutions where some constraints might be violated, and learn about the importance of the constraints from the users' choices between these alternatives. We demonstrate our framework in a user study with a material transport task in an industrial facility. We show that nearly all users accept alternative solutions and thus obtain a revised specification through the learning process, and that the revision leads to a substantial improvement in robot performance. Further, the learning process reduces the variances between the specifications from different users and, thus, makes the specifications more similar. As a result, the users whose initial specifications had the largest impact on performance benefit the most from the interactive learning.*

## 1. Introduction

Mobile autonomous robots are being deployed in a growing number of applications, due to numerous technical advancements in robot capabilities. Despite these advancements in robot autonomy and capability, specifying the robots' task still requires a high level of expertise, which often hinders their acceptance in practice (Villani et al., 2018). Consequently, a key challenge in the field of human–robot interaction (HRI) is enabling a broader range of users to supervise robots.

This requires more than the design of intuitive interfaces for robot programming. Autonomous robots make decisions on their own about how to achieve an objective; supervising and directing their behavior therefore usually demands a deeper understanding of robotics from the user. Thus, new methodologies in HRI that combine accessible interfaces with algorithms to help a novice efficiently use the robot's capabilities are required. This would enable the deployment of robots in more wide-ranging scenarios, especially dynamic settings where robots are required to make autonomous decisions in accordance with human interests.

Research in HRI with non-expert users usually focuses on only one of these two aspects. For instance, Shaikh and Goodrich (2017) and Srinivas et al. (2013) investigated how user interfaces can be made more intuitive. Revising initial specifications was studied for specification languages by Vasile et al. (2017) and Hauser (2014), whereas active preference learning for robotic behavior was discussed by Daniel et al. (2014), Sadigh et al. (2017), and Holladay et al. (2016). Recently, Palan et al. (2019) proposed a combined approach where demonstrations are used as the initialization for preference learning, to speed up convergence.

The novelty of our work is (1) the integration of a specification interface and specification revision via learning, (2)

---

[1]Department of Electrical and Computer Engineering, University of Waterloo, Waterloo, ON, Canada
[2]Monash University, Melbourne, Victoria, Australia

**Corresponding author:**
Nils Wilde, Department of Electrical and Computer Engineering, University of Waterloo, 200 University Avenue West, Waterloo, ON, Canada N2L 3G1.
Email: nwilde@uwaterloo.ca

a learning approach that can obtain improvements with few user interactions on complex tasks, and (3) a validation on a realistic scenario with users.

After users initially provide a specification for how a robot should behave in an environment, we then present them with a visualization of the resulting robot motions, together with alternatives based on a modified specification. Users choose between the alternatives, enabling the learning mechanism to refine their specification. This approach helps especially inexperienced users to deploy robots more efficiently.

We validate the framework in a user study and show that, using our proposed framework, users accept alternative paths and obtain revised specifications, which improve the robot's performance by 14% on average, within at most 20 user interactions. Further, we show that while initially provided specifications vary largely between users, the learning interaction results in specifications that are more similar. We observe that it is especially those users whose constraints initially drastically affect performance who benefit most from the interaction.

We consider an industrial facility where the work space is shared between pedestrians, human-operated vehicles, and autonomous vehicles. While mobile robots are capable of navigating safely given a description of the environment, their choice of routes might not fit the preferences and established rules of humans. Without a further specification, robots are unaware of the context, e.g., areas that are designated for vehicles or areas where robot traffic is undesired. In addition, the behavior of autonomous vehicles can appear unpredictable to humans.

To address these issues, our framework allows users to specify a set of traffic rules to guide robots in such environments. In current industrial practice such rules for robot behavior are designed by trained personnel (OTTO Motors, 2016). We propose an approach for revising specifications through learning to enable inexperienced users to create efficient specifications. In our previous work (Blidaru et al., 2018), we designed a graphical user interface (GUI) where a user can specify traffic rules such as one- and two-way roads, areas of avoidance and reduced speed zones by graphically defining polygons on the map of the environment (we synonymously refer to the user-defined traffic rules as user constraints). However, users of such systems might be oblivious to the impact the specification has on the task performance; strictly following the traffic rules potentially leads to large increases in task completion time. Thus, users might be willing to accept the violation of some of their constraints that are not mandatory, e.g., for safety regulations, and when the violation is sufficiently beneficial for the task performance, i.e., the task completion time. In Wilde et al. (2018), we captured this trade-off by assigning a weight to each constraint, describing the time saving for which violation is acceptable. To reduce the burden on the user, we do not require them to specify these weights, but propose an active learning framework to gain information about the weights through interaction. For a specific task requiring a robot to navigate between a start and goal, we iteratively present the user with two alternative paths the robot could take, illustrated in Figure 1. The user then chooses between these alternatives. Based on the constraints that are violated by the two paths and their respective traversal times, we learn the relative importance of constraints.

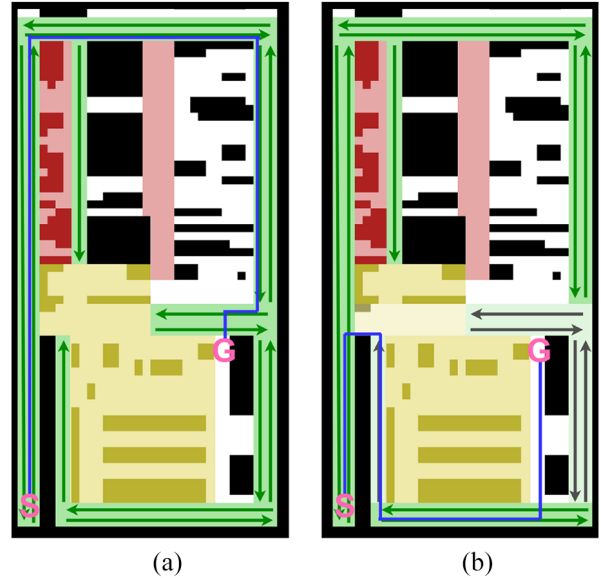

**Fig. 1.** Example environment (white) with obstacles (black), user-defined constraints, and a task start and goal locations. Roads are drawn in green with an arrow indicating the direction. Speed limit zones where only half the maximum speed is allowed are drawn in yellow, while areas of avoidance are illustrated in red. In (a), we see the initial path respecting all user constraints and preferring roads. Following the user interaction, we obtained the revised specification in (b), in which some of the constraints are less important to the user (faded yellow and green). Thus, the shortest path for the given task is significantly shorter, at the cost of a violated speed limit zone, and a violated road zone.

An initial specification assumes that no traffic rules are allowed to be violated. This allows for computing a path that strictly follows the defined rules, shown in Figure 1(a). Through active learning we try to revise the specification and improve the performance. We extend our work in Wilde et al. (2018) to consider multiple start–goal tasks. In each iteration of user interaction, the preferred path becomes the best path so far. If the same task is presented to the user in a later interaction, the previously preferred path constitutes one of the two alternatives. This corresponds to a user-on-the-loop framework: the current best path can already be executed as the user approved it previously. The idea of showing the previously preferred solution in the next iteration is also used in Palan et al. (2019). Depending on the user feedback, the violation of some rules might be acceptable for a certain time benefit. In the example in Figure 1 we show a revised specification following the active learning process, which decreases the task time. Thereby, one speed limit zone has become less important and three roads are rewarded less, such that the robot traverses through the free space, and a one-way road is effectively removed from the specification.

## 1.1. Contributions

We extend our previously introduced active learning system (Wilde et al., 2018) to more complex scenarios consisting of multiple tasks. Thus, the learning algorithm considers two aspects: the task to learn about, and the paths to present to the user for each task. We present an algorithm that evaluates this choice based on the path that results in the largest potential performance increase, and a methodology to combine information learned from user feedback for different tasks. Further, we combine the learning with our GUI from Blidaru et al. (2018) into a framework for robot task specification for inexperienced users. After users initially provide a specification consisting of a set of traffic rules for a mobile robot, their specification is revised using preference learning to yield a more efficient solution. Our second contribution is the evaluation of our framework in a user study. The performance of the learning system was previously demonstrated for single tasks in simulations (Wilde et al., 2018). In this article, we demonstrate the practicality of the multi-task learning system when used by human operators. We show that, given a fixed budget of 20 user interactions, we are able to substantially improve the quality of user specifications. Further, we use the metrics introduced in Blidaru et al. (2018) to systematically evaluate the quality of the specifications provided by users and the revisions obtained through the learning process.

The rest of the article is structured as follows. In Section 2 we review related work before Section 3 introduces the problem statement, briefly reviews our previous work and presents theoretical extensions. Section 4 describes the scenario and procedure of the user study and introduces our main hypotheses, while Section 5 reports the results. We conclude with a discussion and outlook on future work in Section 6.

## 2. Related work

Methods for task specification for autonomous systems can be categorized into three groups: specifications obtained from experts, revision of specifications, and interactive learning.

### 2.1. Specifications obtained from experts

First we review methods for task specification where an expert operator specifies a robot task by either defining reward functions, providing optimal demonstrations or using a specification language. In the first method, reward functions are used to describe the high-level behavior for the robot, which then learns the appropriate policy using reinforcement learning (RL) (Kober et al., 2015; Smart and Pack Kaelbling, 2002). A user-defined reward function maps the system states to a numerical value, expressing how desired that state is. This reward function corresponds to a high-level specification for how the robot should behave; through RL the robot then finds a policy that maximizes the reward. RL has been studied extensively as a tool to realize a high-level description of a robot's behavior (Kober et al., 2015). For instance, Smart and Pack Kaelbling (2002) and Stone et al. (2005) applied RL in the domain of mobile robots and robots competing in soccer games. In both examples the reward function is designed by a human expert. The practicality of RL approaches has also been investigated in field studies (Knox et al., 2013). However, specifying reward functions usually requires a high level of expertise and can be unintuitive.

The field of learning from demonstration (LfD) uses expert demonstrations for robot programming (Argall et al., 2009; Billard et al., 2016). Applications range from high-level task specification (Ekvall and Kragic, 2008) to the definition of precise actions such as grasping (Lin et al., 2012) or manipulation trajectories (Akgun et al., 2012). A common technique in LfD systems is inverse reinforcement learning (IRL) (Abbeel and Ng, 2004). The setting is similar to RL; however, the reward function is unknown. When using IRL in LfD, the objective is to learn how the robot should behave. Demonstrations are provided by a human expert; it is assumed that the human maximizes an internal reward function (Abbeel and Ng, 2004; Ziebart et al., 2008). From multiple demonstrations the learning system tries to recover that reward function in order to imitate the behavior. The reward function is often modeled as a linear combination of pre-defined features; the problem then consists of learning the weights for all features (Ziebart et al., 2008). However, in practice LfD faces challenges when demonstrating the desired behavior requires a high level of expertise (Wilson et al., 2012) or are difficult to provide (Christiano et al., 2017). Specification languages such as linear temporal logic (LTL) (Bhatia et al., 2010) allow for abstract specifications, for instance "*First, visit region A and B, then go to C, and finally visit D.*" In order to reduce the burden on the user, Srinivas et al. (2013) proposed a GUI for LTL mission planning, while Finucane et al. (2010) designed a framework for using natural language to provide LTL specifications.

Our research is closely related to the IRL problem: we want to learn a user's cost function for the constraints they specified, i.e., their importance. Any path that is generated between the start and goal location could be described by a set of features, including those that describe the violation of constraints. Then, learning about the importance of constraints is analogous to recovering a user reward function based on these features.

### 2.2. Revision of specifications

The second approach takes into consideration that demonstrations and specifications, especially when provided prior to the robot executing the task, might be sub-optimal. In the field of LTL, the works of Lahijanian and Kwiatkowska (2016) and Karlsson et al. (2018) both revise an initial specification if it leads to sub-optimal outcomes or is infeasible. In a general motion planning problem on some configuration space with spatial obstacles, Hauser (2014) considered the case when no feasible path exists. The minimal constraint removal problem then finds the biggest subset of obstacles such that a feasibility is re-attained.

The concept of revising initial specifications is also applied to LfD. The work of Niekum et al. (2013) automatically segments the tasks and then efficiently asks for additional demonstrations when needed. Moreover, Grollman and Billard (2011) focused on failed demonstrations. Instead of imitating the human, the learning system tries to avoid repeating the mistakes the operator made.

In a comparable fashion, we receive a set of constraints from a user. We initially set high weights for all constraints such that the resulting path respects all user constraints to yield an initial specification. However, we assume that such a path might not necessarily be optimal as some constraint violations might be allowable. The user agreeing to the violation of a constraint can be thought of as relaxing the constraint in question, which then leads to a revised specification.

### 2.3. Active preference learning

More recently, research has focused on defining the desired behavior of a robot interactively (Christiano et al., 2017; Daniel et al., 2014; Laird et al., 2017; León et al., 2013; Sadigh et al., 2017; Somers and Hollinger, 2016). In active preference learning, users are presented with possible solutions for a defined problem. When they choose between alternatives, the autonomous system learns about their preferences and iteratively improves its strategy. Interactive task specification addresses several drawbacks of the previously discussed techniques. For instance, asking a user for demonstrations is not always desirable, as human demonstrations can be infeasible, e.g., in swarm robotics (Akrour et al., 2012), difficult to provide (Abbeel and Ng, 2004; Daniel et al., 2014), the amount of necessary demonstrations may be prohibitively large (Christiano et al., 2017), or the demonstration itself may require a high level of expertise (Daniel et al., 2014). Providing rich and precise specifications prior to a robot executing a task might also be challenging and more prone to inaccuracies (Abbeel and Ng, 2004). Interactive task specification also improves ease of use by reducing the information required from the user upfront. Instead of asking the user for a complete specification in the beginning or demanding numerous demonstrations, robot tasks can be learned in an iterative, interactive way.

The work of León et al. (2013) and Christiano et al. (2017) addressed these challenges by integrating user feedback into RL systems. Also focusing on RL for autonomous robots, Krening and Feigh (2018) investigated how different interactive learning algorithms are accepted by users and show that users perceive action advice as more effective than action critique in a study with 24 participants. Christiano et al. (2017) applied user interaction to RL. Instead of using human feedback as a reward function user are asked for their pair-wise preference for possible trajectories. This allows to drastically reduce the amount of necessary user interaction.

Recently, numerous contributions to interactive task specification have been made in the field of active preference learning, combining techniques from preference elicitation (Golovin et al., 2010; Guo and Sanner, 2010) and active learning (Jain et al., 2015; Sadigh et al., 2017). The problem of preference elicitation considers a set of hypotheses, tests, and outcomes. By performing tests, some hypotheses become inconsistent with the observed outcomes and are rejected. This can be applied to a robot task specification: hypotheses are possible reward functions of the user. Tests correspond to presenting the user with alternative solutions based on these reward functions, whereas observations are the user's selections. The user's internal reward function is then learned by iteratively ruling out reward functions that become inconsistent with the user's choices.

Active learning allows the learner to decide what query, i.e., what set of alternative solutions the user is presented with next. Daniel et al. (2014) presented a framework where experts rank the performance of a demonstrated grasping task. In Sadigh et al. (2017) and Basu et al. (2018), trajectories for a dynamical system are presented to the user, who then chooses one of two alternatives. Iteratively, weights for trajectory features are learned and an optimal solution is found. Sadigh et al. (2017) validated their work in simulation and in a small user study (10 participants). In both experiments the user model is based on five predefined features. While the simulations demonstrated the convergence of their algorithm over 200 iterations, the user study showed subjective improvements over 10 iterations. The subsequent work of Basu et al. (2018) with richer user feedback is supported by another study with 10 participants that interacted with the learning system for 20 iterations.

Our framework is based on active preference learning. We query the user about their preference for alternative paths and learn about the importance of user constraints from their feedback. However, in our case the set of hypotheses is the set of all possible paths between the start and goal, which is not directly given. When planning on a graph, finding the set of all paths from start to goal is known to be a #P-complete problem (Valiant, 1979). Other work in the field of active preference learning often focuses on user preferences based on robot-centered features, e.g., the work by Sadigh et al. (2017), which are assumed to be known. In our work, we consider environment-centered features that are obtained from an initial user specification. By learning weights for these features we revise the specification. Moreover, in our scenario, we have explicit prior information about the user's preferences. We assume they follow two objectives: minimizing time and only allowing constraint violation when sufficiently beneficial. This may allow us to design strategies for presenting the alternative paths that are either greedily maximizing the potential information gain or that are likely to be accepted by the user.

*2.4. Metrics of user performance*

Finally, our metrics to quantify the impact of user specifications relate to work on measuring the effects of altering a robot's operating environment. Crandall et al. (2005) and Lampe and Chatila (2006) have suggested correlations between the complexity of the robot operating environment and its performance. Multiple methods for measuring complexity have been proposed. In Crandall et al. (2005), Crandall (2003), Crandall and Goodrich (2003), and Crandall and Goodrich (2002), it was suggested that complexity be determined by approximating the branching factor and amount of clutter in the environment. The work of Lampe and Chatila (2006), Yang and Anderson (2011), and Anderson and Yang (2007) proposed a technique rooted in information theory to determine the robot operating environment, with Lampe and Chatila (2006) measuring entropy based on obstacle density, whereas Yang and Anderson (2011) and Anderson and Yang (2007) used the number of accessible neighbors at every location in the motion graph. In addition, Anderson and Yang (2007) proposed a secondary complexity measure based on the distribution of obstacle types and the compressibility of the environment. In Young et al. (2017), the measurement of complexity was extended from a binary distribution of local obstacles to a continuous one, enabling the use of dynamic obstacles.

Our approach uses the work of Anderson and Yang (2007) to compute an entropy-based complexity of the user specifications. The entropy directly correlates with the number of movement-related decisions that the robot will need to make throughout the environment. User specifications tend to decrease the entropy of the environment, and thus increase the predictability of the robot's behavior, by prescribing to the robot how to make decisions. With our focus on warehouse industrial robots, there is a need to ensure an adequate level of predictability in robot behavior, so as to avoid negatively impacting the user's trust and opinion of the robot (Yagoda and Gillan, 2012).

## 3. Proposed approach

*3.1. Preliminaries*

Using definitions from Korte and Vygen (2007), a multigraph is a triple $G = (V, E, \Psi)$, where the function $\Psi : E \to \{(v, w) \in V \times V : v \neq w\}$ associates each edge with an ordered pair of vertices. Given a vertex $v$ we call a vertex $w$ a neighbor of $v$ if $v$ is the start and $w$ the endpoint of an edge in $G$. We denote the set of all neighbors of $v$ as $\mathcal{N}(v)$. A graph is called strongly connected if for any two vertices $v_i, v_j \in V(G)$ there exists a path from $v_i$ to $v_j$ and a path from $v_j$ to $v_i$. Multiple edges are allowed to connect the same ordered pair of vertices and are then called parallel. In our problem we consider doubly weighed multigraphs of the form $G = (V, E, \Psi, c_1, c_2)$, where $c_1$ and $c_2$ are independent weight functions, each associating a real number to each edge of the graph: $c_i : E \to \mathbb{R}$ for $i \in \{1, 2\}$.

A walk between two vertices $v_1$ and $v_{k+1}$ on a graph $G$ is a finite sequence of vertices and edges $v_1, e_1, v_2, e_2, \ldots, e_k, v_{k+1}$ where $e_1, e_2, \ldots, e_k$ are distinct. A path $P_{v_1, v_{k+1}}$ between two vertices $v_1$ and $v_{k+1}$ is defined as a graph $(\{v_1, v_2, \ldots, v_{k+1}\}, \{e_1, e_2, \ldots, e_k\})$ where $v_1, e_1, v_2, e_2, \ldots, e_k, v_{k+1}$ is a walk. On a weighted graph, the cost of a path is defined as $c(P) = \sum_{e \in P} c(e)$. In doubly weighted graphs we define two costs $c_1$ and $c_2$ where $c_1(P) = \sum_{e \in P} c_1(e)$, $c_2(P) = \sum_{e \in P} c_2(e)$.

*3.1.1. Notation.* Vectors are written with bold, lowercase letters, e.g., $\boldsymbol{v}$, we address elements of the vector with a subscript index $v_i$. A superscript index $\boldsymbol{v}^i$ identifies a specific vector. Sets are denoted by uppercase letters ($G$), matrices as bold uppercase letters ($\boldsymbol{A}$).

*3.2. Problem description*

The proposed approach contains the following two components. First, having obtained a user specification, we use an extension of the active learning technique introduced in Wilde et al. (2018), to gain information about the importance of the user constraints, i.e., the user's preference between alternative paths. The technique is extended to allow its use with a multi-task scenario. Following this, we apply the metrics proposed in Blidaru et al. (2018) to evaluate the impact of the specification, and show how the learning system improves the quality of the robot's task performance.

*3.3. Learning user preferences*

*3.3.1. Problem setup.* The learning system receives a description of the environment, the user specification and a set of tasks. The environment is considered to be static and is represented as a weighted strongly connected multigraph $G = (V, E, \Psi, t)$. The weight $t$ on the graph encodes the time a robot requires to traverse an edge. We use parallel edges with different times to model speed. Such a graph can be used by a robot motion planner to navigate through an environment such as that shown in Figure 1. We extend our previous work from Wilde et al. (2018) and consider a set of ordered pairs $\{(s_1, g_1), (s_2, g_2), \ldots\}$ where $s_i$ and $g_i$ are vertices on $G$. A single task consists of navigating from a start $s_i$ to a goal $g_i$. On the environment map, the user specifies a set of constraints $\Gamma = \{\gamma_1, \gamma_2, \ldots, \gamma_d\}$. Each constraint $\gamma_k$ is a pair $(E_k, w_k^*)$, where $E_k$ is a subset of the edges of $G$ and $w_k^*$ is a hidden user weight for the constraint. These weights can be positive or negative: a positive weight $w_k^*$ expresses a penalty for using edges in $E_k$ whereas a negative weight expresses a reward. Note that a road on the interface entails two constraints: a reward for using the road in the direction of travel and a penalty for moving the wrong way. Consequently, a two-way road maps to four constraints.

We incorporate the user specification by creating a doubly weighted graph $G^\Gamma = (V, E, \Psi, t, w^*)$. For each edge $e$ in $G^\Gamma$ the second weight $w^*(e)$ is defined as the sum of all $w_k^*$ that belong to a constraint containing $e$. Our objective is to find paths $P_i^*$ between all $s_i$ and $g_i$ that are optimal with respect to

$$\min_{P_i} \sum_{e \in P_i} w^*(e) + t(e) \quad (1)$$

The true user weights $w_k^*$ are latent, i.e., we do not ask the user to define $w_k^*$ during the specification. Nonetheless, given estimates $\hat{w}_k$ of the true user weights, we can also construct a doubly weighted multigraph $\hat{G}^\Gamma$. Moreover, the weights are defined in units of time, allowing us to pose the multi-objective optimization as an unweighted sum. To learn about the weights, we can query the user. In a query, we present them with a pair of paths $(P_i^1, P_i^2)$ for a selected start–goal pair $(s_i, g_i)$. Considering only pairs instead of more than two paths at a time is motivated by reducing the burden on the user, as choosing between numerous alternatives is more demanding (Jamieson and Nowak, 2011).

*3.3.2. Linear learning model.* Given a specification of $d$ constraints, the latent user weights can be summarized as a column vector $w^* \in \mathbb{R}^d$. Furthermore, a path $P$ is described by the time it takes to traverse $t(P)$ and a vector $\phi \in \mathbb{N}^d$ that indicates for each constraint $\gamma_k \in \Gamma$ how many edges in $E_k$ are traversed by a path, i.e., $\phi_k(P) = |E(P) \cap E_k|$. The cost of a path is then written as $C(P) = \phi(P) \cdot w^* + t(P)$.

We distinguish between *penalty* and *reward* constraints. Penalty constraints include the edges within an avoid zone, edges within a speed-limit zone where the traversal time does not correspond to obeying the speed limit, and the edges going against the defined direction of travel in a one-way road. Reward constraints describe the edges that follow the direction of traffic on a road. Thus, for any penalty constraint the weight $w_i$ is non-negative whereas for a reward constraint the weight is non-positive. As $w^*$ is hidden, we initially only know that its values are finite, i.e., $l_i \leq w_i^* \leq u_i$ for some real number lower and upper bounds $l_i$ and $u_i$.

Let $P^i$ and $P^j$ be two paths. If the user prefers path $P^i$ it implies that $C(P^i) \leq C(P^j)$. We can write this as a half-space in $\mathbb{R}^d$ containing $w^*$

$$\{w \in \mathbb{R}^d | (\phi^i - \phi^j) w^* \leq t^j - t^i\} \quad (2)$$

Thus, obtaining user feedback allows us to iteratively learn inequality constraints on the user weights. We write the intersection of the learned half-spaces as a polyhedron

**Algorithm 1.** $\pi_{\text{vertexSearch}}$, find new weights using DFS.

**Input:** $A, b, k, \mathcal{W}, \hat{w}^{\text{best}}$
**Output:** $\mathcal{W}_{\text{new}}$
1. Initialize set $\mathcal{W}_{\text{new}} = \emptyset$, `openList` $= \{\hat{w}^{\text{best}}\}$ and maximum iterations $i_{\max}$
2. **for** $i = 0$ to $i_{\max}$ **do**
3.   **if** $|\mathcal{W}_{\text{new}}| = k$ *or openList is empty* **then**
4.     **return** $\mathcal{W}_{\text{new}}$
5.   $\tilde{w} = $ `openList.pop()`
6.   **if** $\tilde{w}$ *is not equivalent to any* $\hat{w} \in \mathcal{W}_{\text{new}} \cup \mathcal{W}$ **then**
7.     Add $\tilde{w}$ to $\mathcal{W}_{\text{new}}$
8.   **if** $\tilde{w}$ *is not labelled as discovered* **then**
9.     Label $\tilde{w}$ as discovered
10.   **for** *all* $w' \in$ getAdjacentVertices($\tilde{w}$) **do**
11.     **if** $w' \notin$ `openList` and $\tilde{w}$ is not labelled as discovered **then**
12.       `openList.insert(`$w'$`)`
13. **return** $\mathcal{W}_{\text{new}}$

$$\mathcal{F} = \{w \in \mathbb{R}^d | l_i \leq w_i \leq u_i, \ Aw^* \leq b\} \quad (3)$$

which we refer to as the *feasible space*.

The half-spaces obtained from user feedback can be used to iteratively shrink the feasible space.

*3.3.3. Equivalence regions.* Finally, if a path is optimal with respect to (1) for two different vectors of weights $w^i$ and $w^j$, we call $w^i$ and $w^j$ *equivalent*. This implies that there exist different possible weight configurations that are indistinguishable for the user in our setting, as they correspond to the same path. Consequently, we call a set of weights where all elements are *equivalent* to one another an *equivalence region*. This implies that we do not need to exactly determine $w^*$; it is sufficient to find an estimate $\hat{w}$ that is *equivalent* to $w^*$.

### 3.4. Active learning algorithm

In Wilde et al. (2018), we introduced a learning algorithm based on the notion of equivalent weights and the resulting discretization of the weight space.

In each iteration we pick a weight from the *feasible space* that is not equivalent to the weight of the path that has been presented to the user in the previous iteration. Receiving user feedback allows us to remove at least one equivalence region from the feasible space. As the number of paths and therewith the number of equivalence regions is finite, the algorithm will not be able to find a new weight after a finite number of iterations. Then the algorithm terminates as it has converged to the optimal solution, i.e., all remaining feasible weights are indistinguishable to the user.

Initially, we set the lower and upper bounds for all weights: for penalty constraints, $l_i = 0$, whereas $u_i$ is the sum of all $t_i$ for all edges $e_i$ on the graph $G$. For reward constraints we have $u_i = 0$. However, the lower bound $l_i$ needs to guarantee that there are no negative cycles so that

**Algorithm 2.** $\pi_{\text{minSearch}}$, find minimizing new weights.

**Input:** $A$, $b$, $k$, $\mathcal{W}$, $\hat{w}^{\text{best}}$
**Output:** $\mathcal{W}_{\text{new}}$
1. $\mathcal{W}_{\text{new}} = \emptyset$
2. $\hat{w} = \min \mathbf{1}^T w$ s.t. $Aw \leq b$
3. **if** $\hat{w} \notin \mathcal{W}$ **then**
4. $\quad$ add $\hat{w}$ to $\mathcal{W}_{\text{new}}$
5. **if** $|\mathcal{W}_{\text{new}}| = k$ **then**
6. $\quad$ **return** $\mathcal{W}_{\text{new}}$
7. newSearchDirections $= \emptyset$
8. **for** $\gamma_i$ where $\phi_i(P(\hat{w})) > 0$ **do**
9. $\quad \bar{c} = c$
10. $\quad \bar{c}_i = -1$
11. $\quad$ add $\bar{c}$ to newSearchDirections
12. **for** $\bar{c} \in$ newSearchDirections **do**
13. $\quad \tilde{w} = \min \bar{c}^T w$ s.t. $Aw \leq b$
14. $\quad$ **if** $\tilde{w} \notin \mathcal{W}$ **then**
15. $\quad\quad$ add $\tilde{w}$ to $\mathcal{W}_{\text{new}}$
16. $\quad$ **if** $|\mathcal{W}_{\text{new}}| = k$ **then**
17. $\quad\quad$ **return** $\mathcal{W}_{\text{new}}$
18. $\mathcal{W}' \leftarrow \pi_{\text{vertexSearch}}(A, b, k - |\mathcal{W}_{\text{new}}|, \mathcal{W} \cup \mathcal{W}_{\text{new}}, \hat{w})$
19. **return** $\mathcal{W}_{\text{new}} \cup \mathcal{W}'$

**Algorithm 3.** Choose task for learning.

**Input:** $A$, $b$, $L$
**Output:** $l_{\max}$
1. time_saving $= -\infty$
2. $l^* = \emptyset$
3. **for** $(s_i, g_i, w_i^{\text{best}})$ in $L$ **do**
4. $\quad$ Pick $P^1$ as the optimal path for $\hat{w}_i^{\text{best}}$
5. $\quad w_i^{\text{new}} = \text{minVertex}(A, b, 1, \mathcal{W}_i, \hat{w}_i^{\text{best}})$
6. $\quad$ Compute new path $P^2$ for $w_i^{\text{new}}$
7. $\quad$ **if** $t(P^1) - t(P^2) >$ time_saving **then**
8. $\quad\quad$ time_saving $= t(P^1) - t(P^2)$
9. $\quad\quad l^* = l_i$
10. **return** $l^*$

the optimization problem is well-defined. Let $\gamma_i$ be a constraint containing edges that follow a road. To obtain the lower bound we choose the negative of the length of the shortest edge in the constraint, denoted by $t_i^{\min}$. Further, we subtract a small amount such that a path planner breaks ties in favor of paths using fewer edges: $l_i = -(1 - \epsilon)t_i^{\min}$ where $0 < \epsilon \ll 1$. Then we pick a path $P^0$ that is optimal for some weight $w^0$ where $w_i^0 = u_i$ if $\gamma_i$ is a penalty constraint and $w_i^0 = l_i$ if $\gamma_i$ is a reward constraint. Hence, $P^0$ follows the user specification, i.e., it does not violate any *avoid* or *speed* zones, does not traverse roads in the wrong direction and uses roads as much as possible. In each iteration we then present the user with the current best path and one alternative. If the user prefers the alternative it becomes the new current best path $P^{\text{best}}$.

In Wilde et al. (2018), we presented two policies for finding a new alternative path, $\pi_{\text{vertexSearch}}$ and $\pi_{\text{minVertex}}$, detailed in Algorithms 1 and 2. Both perform a depth first search (DFS) for $k$ steps over the vertices of the current *feasible space*, similar to the pivot step in the simplex algorithm. Furthermore, given a large enough $k$ both policies are guaranteed to find a new path, unless the current feasible space is a subset of an *equivalence region*, in which case the algorithm has converged. While $\pi_{\text{vertexSearch}}$ starts at the weight of the current best path $\hat{w}^{\text{best}}$, its variation $\pi_{\text{minVertex}}$ starts at the minimal weight within the feasible space. This more greedy approach showed a better performance in simulation; hence, we use the minVertex policy in the user study. The corresponding function minVertex(·) takes the following arguments: the current feasible space described by $A$ and $b$, the number of new weights to be returned $k$, the set of weights that were presented in previous iterations $\mathcal{W}_i$ and the current best estimate $\hat{w}_i^{\text{best}}$.

### 3.5. Multiple tasks

We now extend our framework to multiple tasks. The motivation for a multi-task scenario is two-fold. In practice, an autonomous robot usually has to perform more than one task, making the multi-task setting more relevant. Furthermore, learning about several tasks in parallel has potential computational advantages as it allows for an additional degree of freedom in the active learning: choosing what task to learn about in the next iteration. We consider a set of points of interest in the environment yielding multiple start–goal pairs. We learn about the constraints in each interaction round by obtaining feedback for a single task. We can combine the information from multiple rounds by intersecting the *feasible spaces* of all individual tasks. This leads to a *passive* learning effect: obtaining feedback about a task $(s_1, g_1)$ potentially affects the learning for another task $(s_2, g_2)$, as some weights corresponding to paths for $(s_2, g_2)$ might no longer lie in the *feasible* space.

In the multiple task setting we additionally have to pick a start–goal pair for which we want to present new paths. We propose a simple policy for this in Algorithm 3. Let a learning instance $l_i$ be the collection $(s_i, g_i, w_i^{\text{best}})$ for a task $i$ where $w_i^{\text{best}}$ is the weight vector corresponding to the current best path. Further, let $L$ be the set containing all $l_i$ for all tasks in the scenario. Given $L$ and the current feasible space described by $Aw^* \leq b$, the algorithm iterates over all $l_i$ and computes a new alternative path with the minVertex policy (line 4). Then, it selects the task, i.e., start–goal pair, where the time difference between the current best path and the tentative alternative is maximized (line 7). As a result the user is usually presented with those tasks for which the alternatives consist of very different paths in the first few iterations. After some user feedback is obtained, fewer paths are feasible and the respective weights are less different. Hence, in later iterations the two paths presented to the user become more similar.

**Remark.** The evaluation of Algorithm 3 can take significant computation time in practice. We can approximate the selection of $l_i$ by sampling a random subset $L'$ of $L$ and iterate over the elements in $L'$ in line 2 of the algorithm. Further improvements can be achieved by parallel execution of the for loop.

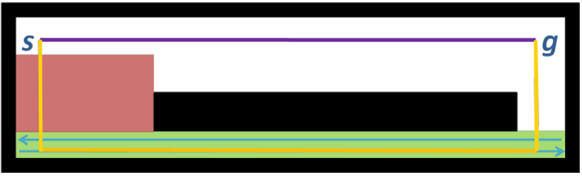

**Fig. 2.** Example for increase in task completion time. The initial specification results in path $P^{\text{init}}$, shown in purple. An alternative solution (yellow) might have a longer traversal time, but correspond better to the user preferences if they value the avoid zone as less important and prefer the use of the road.

*3.5.1. Impact of learning on performance.* Note that the interactive learning does not guarantee improvements in the completion time of paths. Users accept alternative paths if they have a lower cost with respect to (1), which does not necessarily imply a lower time. Consider the simple example in Figure 2. Following the specification leads to the direct path $P^{\text{init}}$, shown in purple. However, a possible alternative enters into the avoid zone (shown in red), but also traverses along a user-specified road (shown in green). The alternative path (shown in yellow) might have a lower cost with respect to the user preference. This effect becomes especially relevant in the multi-task setting due to *passive learning*: obtaining feedback for paths between some $s_1$ and $g_1$ adds inequality constraints to the *feasible space*, which then affects the optimal path between $s_2$ and $g_2$. Relating back to the example from Figure 2, the user might not be presented with these two paths as the learning system might infer about the importance of the avoid zone from a different task. However, in the results of the user study we analyze the performance in detail and show that, in practice, the interactive learning improves the performance for most users.

### 3.6. Metrics

To quantitatively measure the quality of specifications, two metrics are employed: *entropy ratio* and *time ratio*. Similar metrics were originally introduced and applied to user specifications as part of our earlier work in Blidaru et al. (2018).

*3.6.1. Entropy ratio.* Given a graph $G$ and a specification $\Gamma$, the entropy quantifies the complexity of the robot's action space, generated by the combination of the environment and user specification, by considering the number of outgoing edges available at each node, taking their cost into account. The entropy ratio is expressed as the ratio of entropies between graph $G^\Gamma$ and graph $G$ (See Section 3.2), i.e., the constrained and unconstrained environment. We measure complexity using entropy, defined similarly to previous work in Blidaru et al. (2018) and Anderson and Yang (2007). Given an estimate $\hat{w}$ of the user weights, let the cost of an edge be the sum of time and the estimated weight: $\hat{c}(e) = t(e) + \hat{w}(e)$. Further, let $\hat{c}^{\min}(v_i, v_j)$ be the minimal cost between all parallel edges from $v_i$ to $v_j$.

For a given vertex $v_i$ on a graph and the set of its neighbors $\mathcal{N}(v_i)$, the entropy of $v_i$ is given by

$$H(v_i) = -\sum_{v_j \in \mathcal{N}(v_i)} p(v_i, v_j) \log_2 p(v_i, v_j) \quad (4)$$

where we define $p(v_i, v_j)$ as

$$p(v_i, v_j) = \frac{\frac{1}{\hat{c}^{\min}(v_i, v_j)}}{\sum_{k, v_k \in \mathcal{N}(v_i)} \frac{1}{\hat{c}^{\min}(v_i, v_k)}} \quad (5)$$

To obtain the entropy of a graph, we take the sum over the individual vertex entropies:

$$H_G = \sum_{v_i \in V} H(v_i) \quad (6)$$

where $V$ is the set of vertices of a graph $G$, and $H_G$ is the entropy of a graph. The entropy ratio is then denoted by $\eta = \frac{H_{G^\Gamma}}{H_G}$. The entropy is maximized for $H_G$, when there are no user specifications the robot can move freely in any obstacle-free regions of the environment. Adding constraints always decreases entropy as the robot's movement becomes more restricted. Thus, small entropy ratios indicate rigorous specifications where the robot behaves in a more predictable way.

*3.6.2. Time ratio.* Given a graph $G$, a specification $\Gamma$ and a set of start and goal pairs $V'$ where each start and goal is a vertex on $G$, the *time ratio* metric describes the effect of the constraints $\Gamma$ on the average duration of the shortest paths with respect to (1), i.e., the ratio between the average optimal path durations in graph $G^\Gamma$ and in graph $G$. Thereby, paths for all pairs in $V'$ are considered. Similarly to our previous work (Blidaru et al., 2018), we distinguish two forms of the metric: the *global* time ratio considers all vertices on the graph, i.e., $V' = V \times V$ where $V$ is the set of vertices on $G$, whereas the *task* time ratio considers only a defined set of start and goal pairs, i.e., $V' = \{(s_1, g_1), (s_2, g_2), \ldots\}$.

## 4. User study

In this section, we detail the study scenario and procedure and propose our main hypotheses.

### 4.1. Scenario

The study scenario describes a simulated industrial environment adapted from the layout of a real-world facility. An autonomous mobile robot is required to fulfill material transport tasks; a single task consists of navigating from a given start to a given goal location.

Users are provided with a description of the environment detailing different zones as illustrated in Figure 3. This includes areas with high pedestrian traffic, loading docks, storage space, robot parking and charging, dedicated human work and break areas, and an assembly line. The tasks consist of navigating between the labeled areas, for instance traversing from the robot charging zone to the upper end of the assembly line. The user is asked to generate a specification such that the robots are able to reach any of the areas, excluding the human break rooms. Further, robots are never allowed to cross through the assembly line.

Before starting the specification task, users receive an explanation of the traffic rules and instructions on how to use the interface (for more details see Section 4.2). One-way roads are described as encouraging the robot to traverse them in the direction of traffic and discouraging the robot from traversing in the opposite direction. Two-way roads function as two adjacent and opposing one-way roads. Areas of avoidance simply define a part of the environment where robot traffic is undesired, whereas speed limits express that the robot is required to drive with a reduced speed in a specific area.

Finally, users are told that "the robot can navigate freely in the environment without any traffic rules" and has fundamental safety features such as obstacle avoidance. The traffic rules are "meant to guide the robot's behavior" in a way preferable for the user, and users are free to specify as few or many rules as they deem necessary.

### 4.2. Procedure

*4.2.1. Structure.* The study was approved by the Office of Research Ethics at the University of Waterloo. Each study session took approximately 1 hour. The study process is structured in three parts: overview and introduction, training, and main study. In the first part, the scenario and the role of the participant are briefly explained. A video introduces the user interface and demonstrates how to create traffic rules in detail. Further, written information about the traffic rules and the robot's capabilities is provided.

In the training phase the objective is to familiarize participants with the interface. They are presented with a smaller example environment including a similar description as in Figure 3 and are asked to create traffic rules until they feel confident in using the interface. At the end of the training, participants teleoperate the robot in the simulated environment.

The main study has three phases: specification, interaction, and teleoperation. The first two phases are illustrated in Figure 4. In the specification phase, participants are presented with the environment from Figure 3 and the written instructions on the traffic rules and the robot's capabilities. It is once more stated that the robot is required to navigate between all marked areas on the environment (with the exception of the dedicated human break areas) and that the robot is able to navigate without any traffic rules present. Participants are then asked to define the traffic rules they find appropriate to achieve the desired robot behavior.

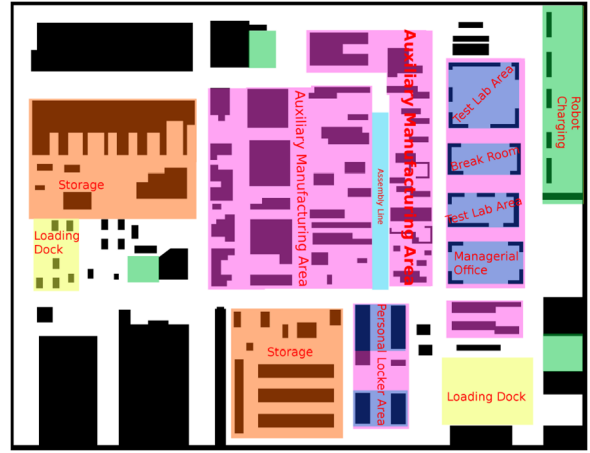

**Fig. 3.** The scenario described in the study. Black corresponds to physical obstacles whereas white is the free space. The described areas in the environment are labeled as follows: high pedestrian traffic, purple; loading docks, yellow; storage, orange; robot parking and charging, green; human work and break areas, dark blue; assembly line, light blue.

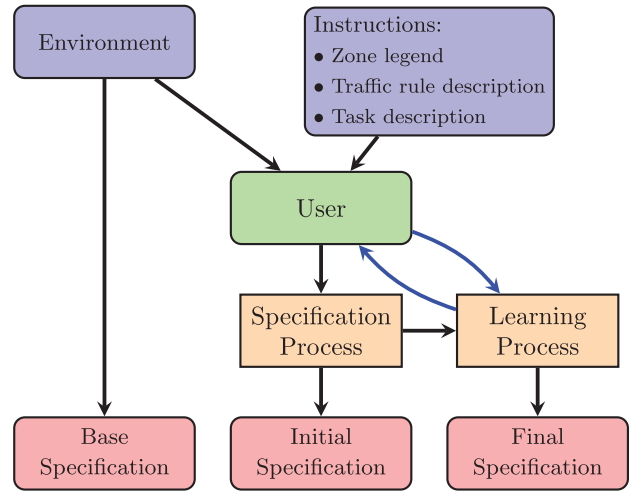

**Fig. 4.** Flowchart of the study with the resulting specifications, black arrows are only executed once whereas blue arrows are executed multiple times. Participants initially receive an instruction set and a description of the environment. The environment yields a base specification, only including obstacles. Using the traffic rules they create the initial user specification for the robot. During the learning interaction users provide feedback, leading to a revised, final specification.

Once they are satisfied with their set of traffic rules, the first phase is concluded.

In the interaction phase, users are iteratively presented with two alternative paths for a task. First, a brief instruction explains the interface: on the map of the environment both paths are shown simultaneously, a simple menu allows participants to select a path, which is then highlighted in color. Further, if a path is highlighted and violates a *penalty* constraint, the perimeter of the constraint in question is also highlighted. Finally, the menu features information about the duration of the two paths and lists the violated penalty constraints. All participants go through 20 iterations of the interaction process, unless the learning algorithm cannot find a new path for any of the tasks and terminates earlier. The task for which they are presented with two alternatives is selected by Algorithm 3 using an approximated set $L'$ containing five tasks, sampled randomly.

In the last phase of the study, participants are asked to teleoperate the robot from a given start to a goal location. Thereby, they can choose freely if they follow their own traffic rules or violate them.

*4.2.2. Questionnaires.* During the study, participants are also presented with several questionnaires. Before providing the specifications, users indicate their trust in the robot's capabilities to fulfill the described tasks. After each of the main steps users are presented with a questionnaire where they rate how well they specified the traffic rules and how confident they feel about using the system. Further, during each step of the interaction participants are asked how acceptable both paths were and, in every third iteration, what their reasoning for their choice was. Finally, the study concluded with a longer questionnaire where users evaluate the overall system. We use the standardized system usability score (SUS) (Bangor et al., 2008) for evaluating the interface design and additional questions focusing on the warehouse scenario.

*4.2.3. Evolution of specifications.* We define the specifications corresponding to different stages of the process, detailed in Figure 4. Before a user specification is provided, the environment including the obstacles yields a base specification where the optimal paths $P_i^{\text{base}}$ for each task only minimize time. After receiving the traffic rules, but before the learning, we obtain the initial specification. The optimal paths $P_i^{\text{init}}$ are the optimal paths corresponding to $w^{\max}$, i.e., the paths that categorically follow the initial specification if possible. After learning, the algorithm has not necessarily converged to the optimal solution due to the limited number of iterations. Therefore, the feasible space might contain weights that are not equivalent to one another. In our learning model, all paths that are optimal for a feasible weight are equally good solutions, so the final path is not uniquely defined. Hence, we need to pick some $w^{\text{final}}$ from the resulting *feasible space* $\mathcal{F}_{\text{final}}$ after learning (see Section 3.3). We propose a conservative approach for determining the final specification by choosing $P_i^{\text{final}}$ to be the optimal paths for the maximum feasible weight, i.e.,
$$w^{\text{final}} = \arg\max_{w \in \mathcal{F}_{\text{final}}}\{\mathbf{1} \cdot w\}.$$

*4.2.4. Types of evaluation.* Further, we categorize different evaluations: interaction-specific evaluation only considers tasks that were presented to the user during interaction. As we directly observed the user choosing between given pairs of paths, we have access to the path characteristics, i.e., time and violation, for each user choice. Task-specific and global evaluations are based on the metrics from the definitions in Section 3.6. Task-specific evaluations consider the shortest paths for all tasks defined in the scenario, whereas global evaluations are based on the shortest paths between all pairs of vertices on the graph. As the interaction is limited to 20 iterations, not every user is necessarily presented with alternative paths for all start–goal pairs. Therefore, task-specific changes between the initial and final specification might result from passive learning.

### 4.3. Interface design

The interface employed in this study is an evolution of that used in Blidaru et al. (2018). It allows users to create a set of robot constraints by defining polygons on the map of the environment. The interface allows for the creation of roads, areas of avoidance and reduced speed zones. In addition, the interface also accommodates the interaction phase, as well as the embedding of study questionnaires that automatically pop-up at predetermined sections of the study. Integrating all the elements of system interaction in a single interface results in a more compact and easy to manage interface, which has been previously correlated to an increase in the resulting human–robot team performance (Steinfeld, 2004; Yanco et al., 2015).

During each iteration of the interaction phase, the interface presents the user with a pair of labeled start and goal points, two alternative paths between these points, the time duration of the two paths, and the constraints that each path violates. The interaction phase interface elements are illustrated in Figure 5. The violated constraints are shown both in list form, highlighted on the map.

In order to measure how users evaluate the system, they are asked several questions throughout the study. To minimize the effect that constant interruptions could potentially have on the users' performance, the questionnaires were integrated directly into the interface in the form of dialog boxes.

### 4.4. Hypotheses

Finally, we propose the two main hypotheses for the user study.

**Hypothesis 1 (H1).** The learning process has the following properties: (a) users accept alternative paths that violate some of the constraints they specified over the course of the learning process and (b) the task performance improves through the interaction process.

**Hypothesis 2 (H2).** Users find the specification process, and the interaction with the learning system intuitive and efficient.

Hypothesis 1 focuses on quantitative analysis of the user interaction. For (a) we analyze the user feedback from the interaction whereas (b) is based on the metrics described in

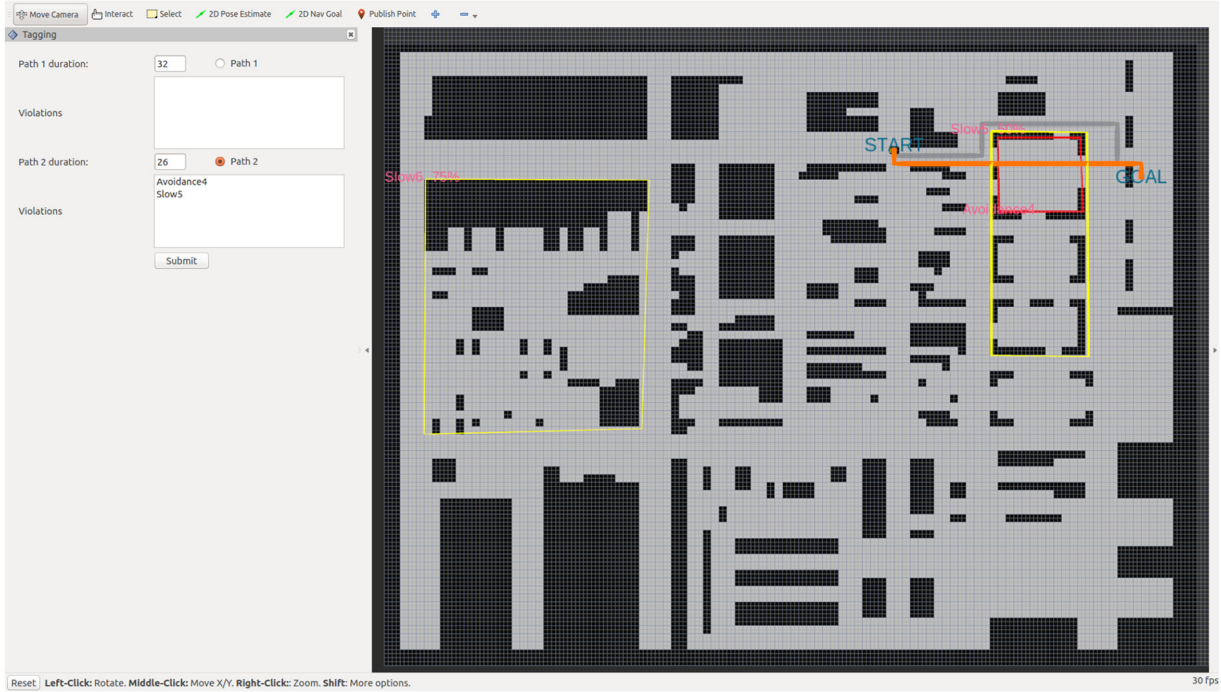

**Fig. 5.** Preference learning user interface displaying duration and traffic rule violations for each path. In addition, the violated traffic rules are also highlighted on the map.

**Table 1.** Characteristics of the initial user specifications. We show the individual mean, median, min, and max for each type of constraint and the number of traffic rules. Further we report the characteristics of the overall smallest and largest specification as well as the example of participant 5 (P 05), shown in Figure 6.

|  | Roads | Avoidance | Speed | Traffic Rules |
| --- | --- | --- | --- | --- |
| Mean | 15 | 4 | 8 | 21 |
| Median | 13 | 5 | 7 | 21 |
| [min , max] | [0, 40] | [1, 9] | [0, 19] | [10, 38] |
| EXAMPLES |  |  |  |  |
| Smallest | 10 | 1 | 5 | 10 |
| Largest | 39 | 1 | 13 | 38 |
| P 05 | 21 | 1 | 17 | 31 |

Section 3.6. To validate Hypothesis 2 we conduct quantitative analyses based on the questionnaire, including the SUS score, as well as a qualitative analysis using the free-form user comments.

## 5. Results

### 5.1. Participants

For the study we recruited 31 participants (21 male and 10 female) via mailing lists. In total 24 participants were affiliated with the Faculty of Engineering at the University of Waterloo. Moreover, 22 participants were pursuing or have completed a graduate education whereas 8 were pursuing or have completed an undergraduate degree. Finally, 6 participants stated that they have background knowledge in robot motion or urban planning.

The population of the study consists of two groups: 21 novice users and 10 repeat users. The novice users had never interacted with the presented framework before, whereas the repeat users had previously used the system once (e.g., during the pilot phase of the study). No participant is part of both groups. In Sections 5.2–5.4 we present results for all users whereas Section 5.5 focuses on differences between the two groups.

### 5.2. Specifications

The initial specifications provided by the users vary in their complexity. We summarize the characteristics of the initial specifications in Table 1. Recall that the number of user-defined roads does not correspond to the number of constraints for the planning problem as roads constitute a reward and a penalty constraint for each lane. We show three example specifications, the smallest and largest with respect to the number of traffic rules as well as the specification from participant 5 that is illustrated in Figure 6.

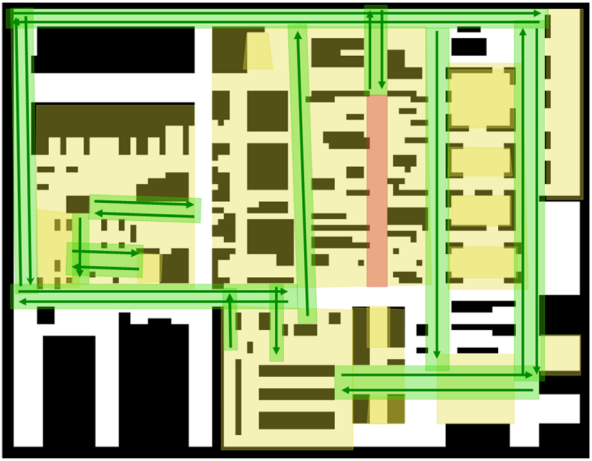

**Fig. 6.** Example specification from participant 5. Reduced speed rules are marked in yellow, road rules in green, and avoidance rules in red.

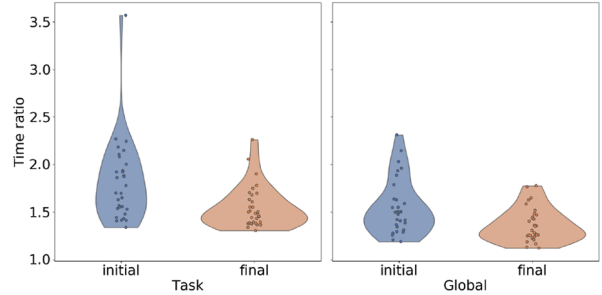

**Fig. 7.** Change in the task-dependent and global time ratio metric of the specification due to active learning. In both plots the left bar shows the time ratio of the initial specification, averaged over all users. The right bar illustrates the time ratio of the final specification, also averaged over all users.

### 5.3. Hypothesis 1

**(a) Acceptance of alternative paths**

For each user we define $\alpha^j_{\texttt{all}}$ to be the percentage of iterations in which user $j$ accepted the alternative path. Further, we introduce $\alpha^j_{\texttt{tasks}}$ as the percentage of the tasks presented to the user where user $j$ accepted at least one alternative, i.e., where they rejected the initial path at some point.

Overall 30 out of the 31 participants accepted at least one alternative path. On average, we found that $\alpha_{\texttt{all}} = 0.44$, meaning that users accepted alternatives in 44% of the interactions. The task-related acceptance has a mean of $\alpha_{\texttt{tasks}} = 0.62$; thus, for roughly 2 out of 3 tasks users preferred an alternative path over the initial one.

Further, we investigate the correlation of $\alpha_{\texttt{tasks}}$ and the richness of user specifications. We characterize the richness of a specification in two ways: the number of traffic rules that the user defined and the number of resulting constraints for the planner. We found that the Spearman rank correlation of the number of traffic rules and the acceptance rate is 0.51 whereas the correlation of constraints and acceptance is 0.60, both with a confidence of $p < 0.005$. This corresponds to a moderate correlation, indicating that users who define a larger set of constraints are more likely to accept alternative paths.

Together with the task-specific acceptance rate of $\alpha_{\texttt{tasks}} = 0.62$ we find strong support for our first hypothesis: users are unaware of the impact of their specification and, thus, allow robots to violate traffic rules (or use roads less frequently) when presented with different possible solutions. Further, the obtained revised specification leads to paths where users chose an alternative path over the initial one in 62% of cases. Moreover, the correlation of complexity and acceptance shows that this effect becomes more apparent for users defining many traffic rules. In Wilde et al. (2018), we postulated three types of users for the simulations: a low-trust user with many constraints for which the importance varies drastically, a high-trust user with few constraints that all are relatively important, and an intermediate user. In the user study we do not observe a discrete separation but rather a continuous distribution for the user behavior. From the difference in the correlation we can conclude that users defining many traffic rules are more likely to accept deviations from the initial path.

**(b) Increased performance**

To evaluate the changes in the performance, we compare the *time ratio* metric of the initial and the final specification, illustrated with violin plots in Figure 7. Further, we compare the metric for global and task-dependent evaluation.

For both evaluations we observe a decrease in the time ratio after the learning process as well as a reduction in the standard deviation. A paired-samples one-sided $t$ test was conducted on the task time ratio between initial and final specifications. The task time ratio of initial specifications ($M = 1.81$, $SD = 0.43$) was found to be significantly different ($p < 0.01$) from that of final specifications ($M = 1.55$, $SD = 0.22$).

Unsurprisingly, the initial specifications vary largely in their impact on the performance as the number of traffic rules users defined range from 10 to 38. However, the decrease in the population standard deviation following interaction indicates that the learning reduces the variation in the performance impact of user input and thus helps users to create more efficient task specifications.

Further, we note that the task-dependent time ratios are higher than the global values for the initial and final specifications. The global metric takes into account locations that are less relevant in the scenario, e.g., the lower left corner of the environment (shown in Figure 6) is not part of a robot task and therefore neglected by most users. Moreover, as the global evaluation considers all vertices on the graph, many close-by pairs of vertices are considered, where the specification often has little influence. Although the task-specific performance is worse, the relative change in time ratio is higher in the task-specific case (14.4%) compared with the global metric (11.8%). Hence, the learning effectively improves the performance of the tasks in the scenario.

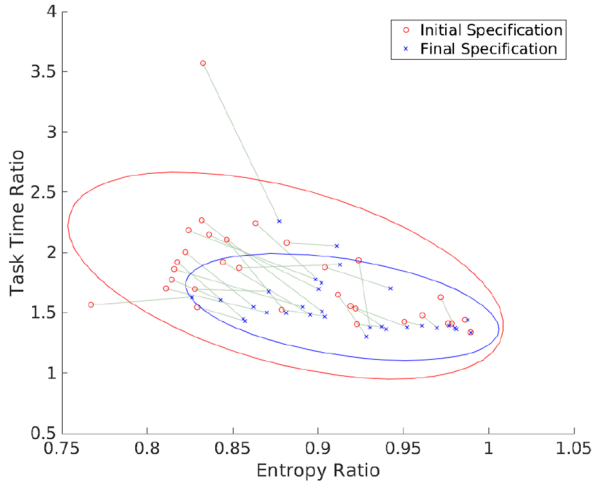

**Fig. 8.** Change in the task time ratio and entropy ratio metrics of the specifications due to active learning. Red indicates the metrics for the initial specification, blue shows the metrics for the final one, and the lines associate the initial and final specifications of each individual users. The ellipses represent the 95% confidence intervals.

Figure 8 illustrates the change of the task-dependent time ratio and the entropy ratio for all user specifications. From the plot, we observe that while the time ratio decreases the entropy ratio increases. Entropy corresponds to how predictably the robot behaves. It captures the robot's degree of freedom with respect to the edge cost on $G^\Gamma$, which is the sum of time and user weight. As the learning process initially assumes high weights on the constraints and thus only reduces weights after obtaining feedback, the entropy increases. After learning, the robot might be allowed to violate some constraints, which enables more options to navigate in the environment. This leads to fewer restrictions on robot behavior, which may not always be desirable. However, this relaxation of the specification is traded-off with the increase in performance.

Running a paired-samples one-sided $t$ test, we found the entropy ratio of initial specifications ($M = 0.881$, $SD = 0.064$) to be significantly different ($p < 0.01$) from the entropy ratio of final specifications ($M = 0.9142$, $SD = 0.046$). Moreover, we notice that while the mean entropy increases, the standard deviations of the time and entropy ratios decrease due to the learning. With respect to the metrics, the specifications become more similar during the learning. Two-sample $f$ tests between the initial and final time ratios show a significant difference ($p < 0.01$) in the variances, both in the case of global and task versions of the metric. However, no significant difference in the variances was found when performing the two-sample $f$ test between the initial and final entropy ratios of the specifications. In addition, Figure 8 also suggests that the specifications with low initial values of entropy and high initial task time ratios generally see more improvement following the preference learning process. This is verified by a strong Pearson correlation between the initial values and the difference between the final and initial values, resulting in $\rho = -0.88$ ($p < 0.01$) for time ratio, and $\rho = -0.85$ ($p < 0.01$) in the case of entropy ratio.

In summary, the learning system leads to a significant improvement in the time ratio metric, especially when measured for the tasks in the scenario. Further, the learning revision reduces the variance between the performance of specifications. Moreover, specifications that are initially more inefficient benefit more from the learning process.

### 5.4. Hypothesis 2

We now report on the users' assessment of the usability of our framework, based on the SUS. Although the SUS does not provide a grade for the usability itself, the work of Bangor et al. (2008) provides a reference frame based on 2,324 surveys using the SUS. In our study, users gave a mean SUS score of 69 whereas the median is 75. The difference arises from two outliers in the dataset with a difference from the mean of over 2 and over 3 standard deviations. The mean corresponds to the second highest quarter of all surveys examined in Bangor et al. (2008). Specifically, for computer-based GUIs, Bangor et al. (2008) reported a mean of SUS of 75.

After the three main parts of the study, constraint specification, learning interaction, and teleoperation, participants were asked to asses how well they specified the robot behavior on a 1 to 10 scale. On average, users reported similar ratings at each step, varying between 7.5 and 7.9, with standard deviations between 1.1 and 1.6. Hence, users felt relatively confident about how they used the framework. Interestingly, we observe an inverse correlation (Spearman coefficient $-0.65$, $p < 0.01$) between the second self-assessment and the richness of the specification, i.e., the number of constraints. Users defining a larger set of constraints tended to view their specification more critically after the learning. Thus, the interactive framework helps users to better understand the impact of their specification on the robot's performance.

### 5.5. Differences in the population

*5.5.1. Acceptance rates.* When splitting the data into the two populations, we observe only a minor increase in the acceptance rates for the novice users compared with the repeat ones. However, the correlation of acceptance rate and complexity of the specifications disappears for the repeat users whereas it is stronger for novices. Repeat users are more aware of the impact of their specifications whereas novice users benefit from the interaction to improve the robot's behavior.

*5.5.2, Time ratio metric.* Between novice and repeat users the *time ratio* metric varies. We recall that the task-dependent metric better reflects the effect of specifications on the task performance and, thus, we show the results for the task-dependent time ratio in Figure 9. We observe that the initial specifications provided by the novice users show a larger variance compared with repeat users. While the median values are relatively similar, the distribution for novice specifications spreads out to higher time ratios. However, the time ratios of final specifications are much more similar.

A two-sample one-tailed *t* test was performed on the difference in time ratio between the initial and final specifications of novice and repeat users, which revealed that the two populations are significantly different ($p < 0.01$). In other words, the changes in the time ratios differ between the two groups.

We conclude that novice users create more diverse specifications with respect to the impact on performance. However, the learning process helps them to improve the specification and obtain better performance. Repeat users seem to have a better understanding of the effect of the traffic rules and, thus, design specifications more carefully. Consequently, they allow for fewer violations that effectively render constraints insignificant and therefore obtain a smaller time benefit.

*5.5.3. Entropy ratio.* Although small differences in the mean entropy ratios across the two populations were observed for both the initial (0.890 for novice and 0.870 for repeat) and final (0.922 for novice and 0.901 for repeat) specifications, these differences were not found to be statistically significant.

*5.5.4. SUS score.* The mean SUS of repeat users is 70 (median 77) whereas the mean of novice users is 68 (median 74). Naturally, participants who have interacted with the GUI before are likely to find it more easy to use. Nonetheless, the reported difference is less than half of the standard deviation among all users scores and thus is not statistically significant. This supports our claim that the presented framework is suitable for inexperienced users.

## 6. Discussion and future work

### 6.1. Summary

In this article, we have presented a framework to allow users to specify spatial and temporal constraints, i.e., traffic rules, for robot movement on a map. This yields an initial specification where each path a robot would take to accomplish a task follows all constraints. However, users might misjudge the impact of their constraints on task performance as well as robot behavior and, thus, may allow the violation of constraints to enable the robot to take more preferable paths. Therefore, the initial specification is revised using active preference learning: the user is iteratively presented with the current path for some task and an alternative solution. From their feedback the system learns about the importance of constraints which we represent by weights. After up to 20 iterations of user interaction, a revised specification is obtained. In the study we observed that all but one user accepted alternative paths during the interaction and we improve the task-specific performance on average by 14%. Further, users were generally positive regarding the usability of the GUI. In this section, we discuss additional findings in the study that do not directly relate to our introductory claims or the proposed hypotheses.

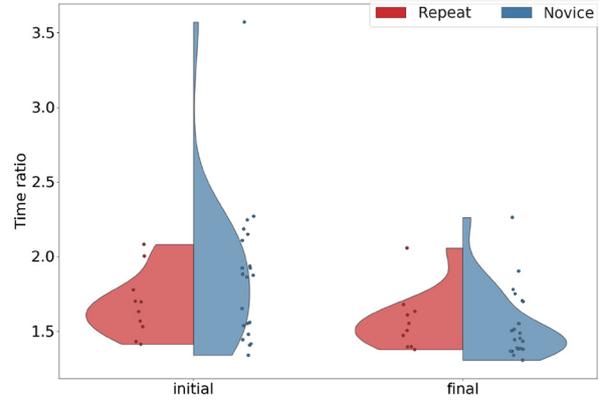

**Fig. 9.** Change in the task-dependent time ratios of the specification, comparing novice and repeat users.

### 6.2. User feedback

Although most users ranked the interaction with our system as positive, participants provided several suggestions for improving the framework in the questionnaire feedback. Almost half the users expressed the desire to change their specification during the learning process. This indicates that although instructed on general robot behavior, participants found it somewhat difficult to envision how all of the created traffic rules affect the behavior of the robot. They could be well served by visualizations showing robot behavior during the specification phase.

Another aspect that could be improved is how the user feedback is incorporated into the learning. Currently users express their traffic rules preferences by selecting the preferred path. Although this approach is intuitive and simple to use, users occasionally expressed frustration when both paths presented to a user contain undesirable behavior, and so users have to select the lesser of two evils. As a result, future work should investigate additional forms of feedback that might better reflect a user's preference, and could potentially lead to a more efficient learning process.

The work of Basu et al. (2018) investigated richer forms of feedback in active preference learning. In addition to ask for the user's preference, feature queries give the user the opportunity to express their reasoning, i.e., "Which feature is most responsible for the difference in your preference between these two trajectories?" This aligns with feedback from the questionnaire: some participants stated that they rejected alternative paths as they violated both a minor and a major constraint at the same time; the violation of only the minor constraint would have been acceptable, but that was unknown to the learning system. In this case richer user feedback would help in two ways, allowing users to express the reasoning for their path selection, and potentially reducing the number of iterations of the learning. On the other hand, a drawback of this approach is the increased complexity of the interaction.

Another approach for richer feedback could allow users to manually indicate, and potentially correct, the undesirable sections of presented paths. This idea is investigated by Cui and Niekum (2018) where users segment a robots trajectory into good and bad parts.

### 6.3. Repeat and novel users

In Section 5.5, we have shown that specifications originating from novice and repeat users differ in the time ratio metric. This indicates that there are differences in how the two groups of users specify the robot constraints, and that these metrics could be used to identify the expertise of a user based on the specification provided. In multi-user systems, this could be used to combine multiple specifications, emphasizing those of expert users. Despite the observed differences, the iterative preference learning system was shown to be capable of improving the specification performance of both types of users. This leads us to hypothesize that even in the case of specifications designed by domain experts, the learning framework could still be used to help increase specification performance. To further investigate this, we currently develop a web-based user interface for remote study participation, allowing us to recruit more domain experts, i.e., people working in industrial facilities.

### 6.4. Learning framework

In our previous work Wilde et al. (2018), we evaluated the active learning framework in simulation. Validating the extended algorithm proposed here in the user study allows us to make additional observations about the practicality of the approach. Unlike the work of Sadigh et al. (2017), Daniel et al. (2014), Guo and Sanner (2010), and Golovin et al. (2010), our learning framework is currently based on a deterministic user model. The major drawback is that our model does not consider users who behave differently than described in the assumed cost function. Nonetheless, we were able to demonstrate that using a simplified linear user model, the framework proposes alternative paths that users accept over the initial paths and revises the specification to improve the task performance within a small number of iterations. Although the resulting final specification does not necessarily correspond to the optimal solution with respect to the hidden user preferences, the deterministic model allows for quick learning, yielding substantial improvements within only 20 iterations. A more complex, potentially probabilistic user model would make fewer assumptions about the user's behavior and, thus, be more robust; however, usually at the cost of performance, i.e., the number of iterations required for learning in a comparable setting.

Further, due to the multi-task scenario we were able to observe some inaccuracies in the user feedback with respect to our user model. When learning about a single task, the *feasible space* can never be empty as the algorithm stops when all feasible weights are equivalent. However, intersecting the feasible spaces for different tasks can lead to an empty set. In that case, the user feedback to different tasks contradicts one another, assuming the linear cost function. Note that an empty intersection of the feasible spaces is not a necessary but a sufficient condition for inaccurate user feedback. In the study we observe this phenomenon for a total of 4 out of the 31 users. A more expressive user model could be used to avoid this issue of converging to a suboptimal solution, however, a richer model is likely detrimental to the efficiency of the learning.

In summary, the user study successfully validates the active learning framework and helps users to create better specifications. The algorithm assumes users' preferences can be described by a deterministic linear model, which should be addressed in future work, for instance by using a probabilistic model for the user. In Wilde et al. (2019), we proposed such a model together with an adapted learning algorithm and demonstrated performance and robustness in simulations. Nonetheless, the simpler user model is beneficial for performance; the study showed that the current algorithm allows for a substantial improvement of specifications while requiring few iterations of user interaction. In particular, users who generate initial specifications most detrimental to robot performance received the most benefit from the interaction, resulting in final specifications that are more similar across users, and thus reducing the need for user training.

**Funding**

This research was partially supported by the Natural Sciences and Engineering Research Council of Canada (NSERC) and OTTO Motors.

**ORCID iD**

Nils Wilde  https://orcid.org/0000-0003-3238-8153